\documentclass[letterpaper]{article} 
\usepackage{aaai24}  
\usepackage{times}  
\usepackage{helvet}  
\usepackage{courier}  
\usepackage[hyphens]{url}  
\usepackage{graphicx} 
\usepackage{natbib}  
\usepackage{caption} 
\usepackage{algorithm}
\usepackage{algorithmic}

\usepackage{amsthm}
\usepackage{amsmath}

\usepackage{enumitem}
\usepackage{booktabs}
\usepackage{xcolor}
\definecolor{mygrey}{cmyk}{0, 0, 0, 59}

\usepackage{multirow}
\usepackage{amssymb}
\usepackage{bm}
\usepackage{colortbl}
\usepackage{tabularx}
\newcommand*\samethanks[1][\value{footnote}]{\footnotemark[#1]}

\usepackage{newfloat}
\usepackage{listings}
\DeclareCaptionStyle{ruled}{labelfont=normalfont,labelsep=colon,strut=off} 
\floatstyle{ruled}
\newfloat{listing}{tb}{lst}{}
\floatname{listing}{Listing}
\title{Prompt-based Distribution Alignment for Unsupervised Domain Adaptation}
\author{
    Shuanghao Bai\textsuperscript{\rm 1},
    Min Zhang\textsuperscript{\rm 2},
    Wanqi Zhou\textsuperscript{\rm 1,\rm 4},
    Siteng Huang\textsuperscript{\rm 2},
    Zhirong Luan\textsuperscript{\rm 3}, \\
    Donglin Wang\textsuperscript{\rm 2\thanks{Corresponding authors.}} ,
    Badong Chen\textsuperscript{\rm 1\samethanks} 
}
\affiliations{
    \textsuperscript{\rm 1}Institute of Artificial Intelligence and Robotics, Xi'an Jiaotong University, Xi'an, China \\
    \textsuperscript{\rm 2}Westlake University
    Institute of Advanced Technology, Westlake Institute for Advanced Study \\
    \textsuperscript{\rm 3}School of Electrical Engineering, Xi’an University of Technology, Xi'an, China
    \textsuperscript{\rm 4}RIKEN AIP
}

\usepackage{bibentry}

\begin{document}

\maketitle

\begin{abstract}
\label{sec:abs}

Recently, despite the unprecedented success of large pre-trained visual-language models (VLMs) on a wide range of downstream tasks, the real-world unsupervised domain adaptation (UDA) problem is still not well explored.
Therefore, in this paper, we first experimentally demonstrate that the unsupervised-trained VLMs can significantly reduce the distribution discrepancy between source and target domains, thereby improving the performance of UDA. 
However, a major challenge for directly deploying such models on downstream UDA tasks is prompt engineering, which requires aligning the domain knowledge of source and target domains, since the performance of UDA is severely influenced by a good domain-invariant representation.
We further propose a \textbf{P}rompt-based \textbf{D}istribution \textbf{A}lignment (\textbf{PDA}) method to incorporate the domain knowledge into prompt learning. Specifically, PDA employs a two-branch prompt-tuning paradigm, namely base branch and alignment branch.
The base branch focuses on integrating class-related representation into prompts, ensuring discrimination among different classes. 
To further minimize domain discrepancy, for the alignment branch, we construct feature banks for both the source and target domains and propose image-guided feature tuning (IFT) to make the input attend to feature banks, which effectively integrates self-enhanced and cross-domain features into the model. 
In this way, these two branches can be mutually promoted to enhance the adaptation of VLMs for UDA.
We conduct extensive experiments on three benchmarks to demonstrate that our proposed PDA achieves state-of-the-art performance. The code is available at https://github.com/BaiShuanghao/Prompt-based-Distribution-Alignment.

\end{abstract}

\section{Introduction}
\label{sec:intro}

\begin{figure}[ht]
  \centering
  \includegraphics[width=0.472\textwidth]{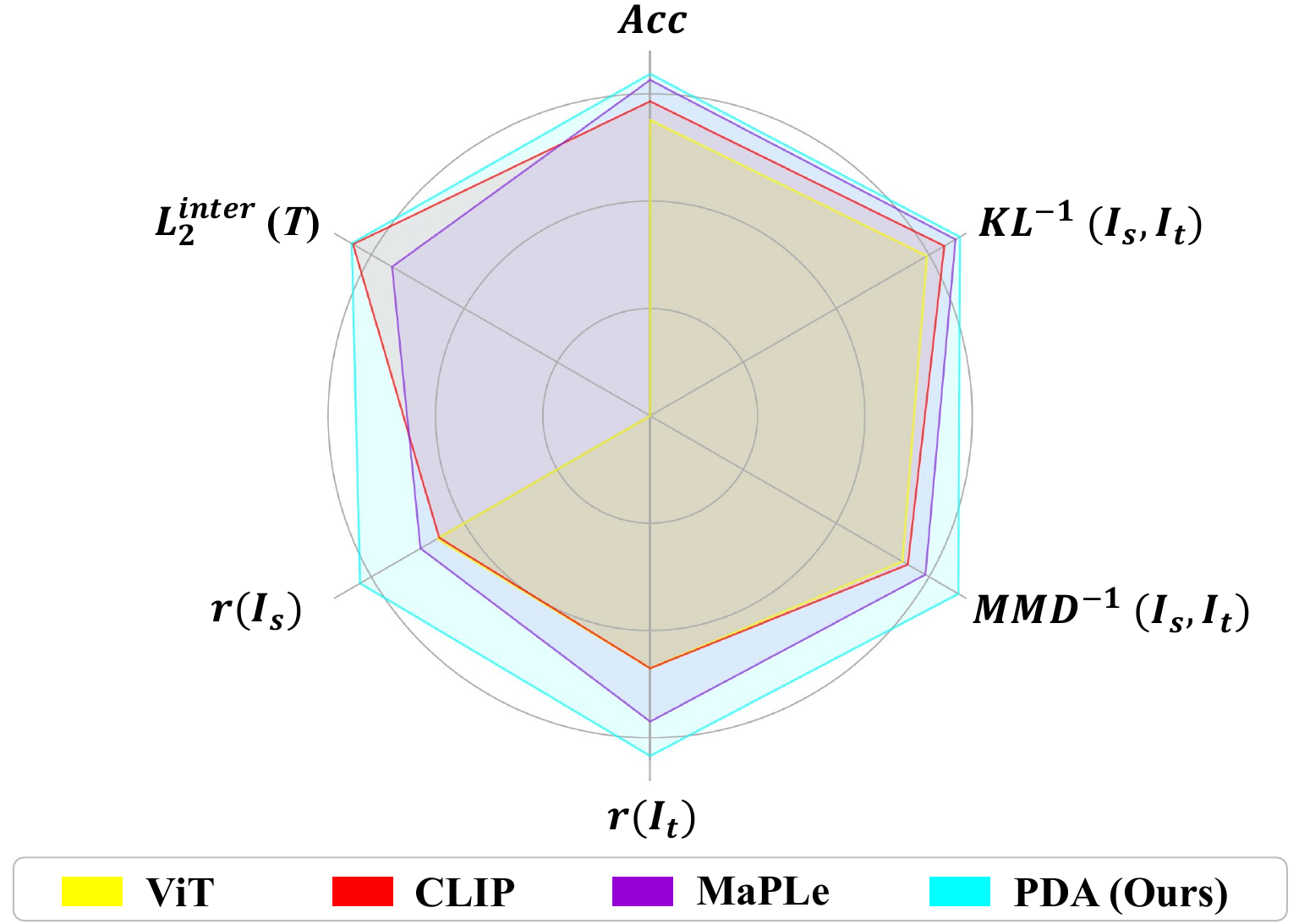}
  \caption{Metric comparisons on Office-Home. Higher values are better. $r$ measures the compactness of features~(\textit{i.e.}, the division of inner-class $L_2$ distance and inter-class $L_2$ distance $L^{inter}_2$). MMD and KL divergence measure the domain discrepancy. $T$, $I_s$ and $I_t$ denote the text features, and image features of the source and target domain, respectively. Our method demonstrates the most discriminable text features, the most compact image features, the lowest domain discrepancy, and the best accuracy.}
  \label{fig:metric}
\end{figure}

Unsupervised domain adaptation (UDA) aims to improve the generalization performance in the target domain of the pre-trained model by using the labeled source domain and unlabeled target domain~\cite{wilson2020survey,zhu2023generalized}.
Many methods have been proposed to address the UDA problem, mainly including  
adversarial training~\cite{ganin2015unsupervised, rangwani2022closer} and metric learning~\cite{saito2018maximum, tang2020unsupervised,zhang2020knowledge}.
However, mitigating distribution by domain alignment may inadvertently result in a loss of semantic information, which comes from the entangled nature of semantic and domain information~\cite{tang2020unsupervised, ge2022domain,zhang2022domain}.

Recently, large vision language models~(VLMs) like CLIP~\cite{radford2021learning} have shown impressive generalization performance in various downstream tasks. With the disentangled visual and semantic representations, this may avoid the loss of semantic information and improve UDA performance. 
In light of this, we conduct an empirical experiment to demonstrate the applicability of VLMs to the UDA problem. Specifically, we evaluated the performance of both unimodal model Vision Transformer~(ViT)~\cite{dosovitskiy2020image} and zero-shot CLIP with hand-crafted prompts.
In Figure~\ref{fig:metric}, although the compactness of source features $r(I_s)$ and target features $r(I_t)$ of CLIP is similar to that of supervised-trained ViT, yet maximum mean discrepancy (MMD) and KL divergence (KL) minimize, resulting higher accuracy of target domain (Acc). This indicates CLIP has the potential to minimize the domain discrepancy for UDA, which benefits from the multi-modal interaction.

To further adapt VLMs to downstream UDA tasks, one of the most efficient paradigms is prompt tuning. Current state-of-the-art prompt tuning methods, such as CoOp~\cite{zhou2022learning} and MaPLe~\cite{khattak2023maple}, have demonstrated superior performance on some specific downstream tasks. CoOp method adopts soft prompts to learn an appropriate text prompt, and MaPLe further introduces vision-language prompts to ensure mutual synergy.
As shown in Figure~\ref{fig:metric}, we observe that 1) MaPLe takes a step towards aligning domains compared to CLIP, as evidenced by its lower KL divergence and MMD, which indicates that \textit{the prompts tuning can help minimize the domain shift.} 2) The image features of MaPLe are more compact, indicating \textit{prompt tuning can further improve the discriminative ability of CLIP model.}
Nonetheless, these prompt tuning methods such as CoOp or MaPLe may not be sufficient to address the domain shift problem fully because these methods primarily focus on the placement of the prompt and may not directly tackle the underlying causes of the domain shift. 
Therefore, we argue that prompts should not only focus on their design but also adapt to different domains by incorporating domain knowledge into the prompt.

To this end, we propose a \textbf{P}rompt-based \textbf{D}istribution \textbf{A}lignment (\textbf{PDA})
method for UDA. PDA consists of two branches, namely the base branch and the alignment branch. 
The base branch generates the image and text representations with prompt tuning, which focuses on integrating class-related representations into prompts, ensuring discrimination among different classes for each domain. 
The principal objective for UDA is to minimize the distribution shift of image representations. The alignment branch utilizes image representations to introduce domain knowledge to minimize the domain discrepancy. To achieve this, we first construct a source-domain and target-domain feature bank and propose image-guided feature tuning (IFT) to make the image representations of inputs attend to feature banks, which can effectively integrate self-enhanced and cross-domain features into the model. 
As shown in Figure \ref{fig:metric}, PDA not only excels in obtaining more discriminable image and text representations but also effectively mitigates the domain discrepancy.
Therefore, our method can guarantee the discriminability of the model, and effectively capture important features from both the source and target domains, which enables domain alignment and allows the model to better adapt to the target domain. 
Our main contributions are as follows:

\begin{itemize}
    \item We first experimentally verify the effectiveness of VLM on UDA downstream tasks. Then, based on this finding, we further propose a prompt-based distribution alignment (PDA) method to tune prompt to the target domain.
    \item The proposed PDA includes two training branches. First, the base branch ensures discrimination among different classes. Second, the aligned branch obtains the domain-invariant information by image-guided feature tuning. 
    \item Extensive experiments demonstrate the effectiveness of the proposed PDA, which achieves state-of-the-art performance on Office-Home, Office-31 and VisDA-2017.
\end{itemize}

\section{Related Work}
\label{sec:rw}

\subsection{Unsupervised Domain Adaptation}

Unsupervised domain adaptation~(UDA) aims to align the source and target domains by learning a domain-invariant feature representation~\cite{zhang2023rotogbml,chen2022ba,xiao2022decoupled}. One method of aligning domains is minimizing divergence between different domains. Many divergence measures have been proposed, such as maximum mean discrepancy (MMD)~\cite{long2015learning}, correlation alignment (CORAL)~\cite{sun2016return} and maximum density divergence (MDD)~\cite{zhang2019bridging}. 
Another line of work is motivated by the success of adversarial learning. By modeling the optimization process as a minimax problem~\cite{ganin2015unsupervised, long2018conditional, rangwani2022closer, xiao2021learning}, a domain discriminator is introduced to distinguish the samples from different domains, with the aim of training the model to generate domain-invariant features that can deceive the domain discriminator. 
With the advent of transformer models, TVT~\cite{yang2023tvt} proposes an adaptation module to obtain both transferable and discriminative features, and CDTrans~\cite{xu2021cdtrans} leverages the robustness of cross-attention modules and proposes a cross-domain transformer for direct feature alignment. 
Different from these mainstream unimodal UDA methods, we focus on harnessing the transferability inherent in vision language models, which exhibit a promising capacity for domain alignment due to multimodal interaction.

\subsection{Vision Language Models}

The pre-trained Vision Language Models (VLMs) learn image-text correlation by various pre-training tasks, such as masked language modeling ~\cite{kim2021vilt}, masked language modeling ~\cite{tan2019lxmert}, image-text matching ~\cite{huang2021seeing} and contrastive learning ~\cite{jia2021scaling, zhang2022tree, chen2021pareto}. 
Although these models have achieved unprecedented success across a wide range of tasks including zero-hot and few-shot visual recognition, effectively adapting them to downstream tasks remains a formidable challenge. 
Many works have been proposed to enhance the generalization ability on downstream tasks by introducing additional feature adapter ~\cite{gao2021clip, zhang2023map, bai2024improve}, attention ~\cite{guo2023calip}, cache model ~\cite{zhang2022tip} and so on. 
The prompt learning paradigm, initially employed in the field of Natural Language Processing (NLP), has also been integrated into VLMs, emerging as one of the most efficient approaches for fine-tuning VLMs on various downstream tasks.
In this work, we follow the line of prompt learning methods and propose a prompt-based distribution alignment method to improve the transferability of CLIP for addressing the UDA problem.

\subsection{Prompt Tuning in Vision Language Models}

Prompt tuning is one of the important parts of parameter-efficient tuning, which aims at learning only a small number of parameters by means of input composition ~\cite{pfeiffer2023modulardeeplearning, zhu2023bridging} while keeping the large model fixed. 
CoOp ~\cite{zhou2022learning} firstly introduces soft prompt in VLMs, demonstrating that suitable text prompts can enhance image recognition performance. 
CoCoOp ~\cite{zhou2022conditional} extends the CoOp by integrating lightweight neural networks to dynamically generate prompts for individual images to deal with the overfitting problem of prompts.
VPT ~\cite{jia2022visual} achieves impressive results using a few visual prompts in transformer models.
Furthermore, MaPLe ~\cite{khattak2023maple} combines both text and visual prompts into CLIP to improve the alignment between text and image representations.
To exploit the effectiveness of prompt tuning for UDA, we introduce a two-branch training paradigm consisting of base and alignment branches. The base branch leverages prompt tuning to enhance the discriminability of CLIP model. For the alignment branch, we design an image-guided feature tuning to mitigate domain discrepancy.

\section{Preliminaries}
\label{sec:pre}

\subsection{Unsupervised Domain Adaptation}

UDA focuses on improving the model's generalization performance with the labeled data from the source domain and unlabeled data from the target domain. Formally, given a labeled dataset $D_s=\{{x^{s}_{i}, y^{s}_{i}}\}^{n_s}_{i=1}$ of the source domain and unlabeled dataset $D_t=\{x^{t}_{j}\}^{n_t}_{j=1}$, where $n_s$ and $n_t$ denote the size of samples in the source and target domains, respectively. 
Note that the data of two domains are sampled from two different distributions, and we assume that the two domains share the same label space.
We denote the input space as $X$ and denote the label set as $Y$. There is a mapping $M: \{{X}\} \rightarrow Y$ from images to labels. 
In this work, we incorporate prompts $V$ into the input, thus the mapping could be rephrased as $M: \{{X, V}\} \rightarrow Y$ from images and prompts to labels. 
Our goal is to mitigate the issue of domain discrepancy between $D_s$ and $D_t$, and to learn a generalized prompt $P$ that can facilitate the transfer of knowledge from the source domain to the target domain. 

\subsection{Revisiting Prompt Learning}

Contrastive Language-Image Pre-Training (CLIP) model consists of an image encoder and a text encoder, which encodes images and corresponding natural language descriptions, respectively.

\noindent \textbf{Zero-shot inference.} The pre-trained CLIP model is adapted to downstream tasks with hand-crafted prompts, rather than fine-tuning the model. 
The text is always manually designed as "a photo of a [CLASS]"~([CLASS] is the class token). 
The image-text matching score is computed using the cosine similarity $sim(w_i, z)$ between the image representation $z$ and the text representation $w_i$ corresponding to the i-th class. The image representation is derived from the image encoder with an input image, while the text representation $w_i$ is extracted from the text encoder using the prompt description associated with the i-th class.
The probability of the image belonging to the i-th class can be formulated as:

\begin{equation}\label{func:zs}
p(y=i \mid x)=\frac
{\exp \left(sim\left(w_i, z)\right ) / t \right)}
{\sum^K_{j=1} \exp \left(sim\left(w_j, z)\right ) / t \right)},
\end{equation}
where $t$ denotes temperature parameter, $K$ denotes the number of classes and $sim$ denotes the cosine similarity.

\noindent \textbf{Text prompt tuning.} It avoids prompt engineering manually and strengthens the transferring ability of CLIP. CoOp ~\cite{zhou2022learning} introduces a set of $M$ continuous learnable context vectors $v=[v^1,v^2,...,v^M]$, then the i-th class of text prompt ${t}^i$ is defined as ${t}^i=[v,c^i]$, where $c^i$ is the fixed input token embedding. The learnable context vectors can be extended to deeper transformer layers of the text encoder with transformer-based architecture, thus each layer of input can be rephrased as $[v_j,c_j]^J_{j=1}$, where $J$ is the number of transformer layers in the text encoder and $[\cdot,\cdot]$ refers to the concatenation operation.

\noindent \textbf{Visual prompt tuning.} It adopts a similar paradigm as text prompt tuning, where additional context vectors that are fed into each layer of the image encoder are automatically learned. For transformer-based image encoder, VPT ~\cite{jia2022visual} inserts a collection of prompts $\widetilde{v}$ between a sequence of patch embeddings $e$ and the learnable class token $c$, which can be designed as $[\widetilde{v}_j,e_j,c_j]^J_{j=1}$.

\noindent \textbf{Multi-modal prompt tuning.} The text prompt $v$ and visual prompt $\widetilde{v}$ are combined into CLIP. For instance, MaPLe ~\cite{khattak2023maple} tunes the vision and language branches of CLIP together by sharing prompts across both modalities.

\section{Method}
\label{sec:method}

\begin{figure*}[htbp]
  \centering
  \includegraphics[width=\textwidth]{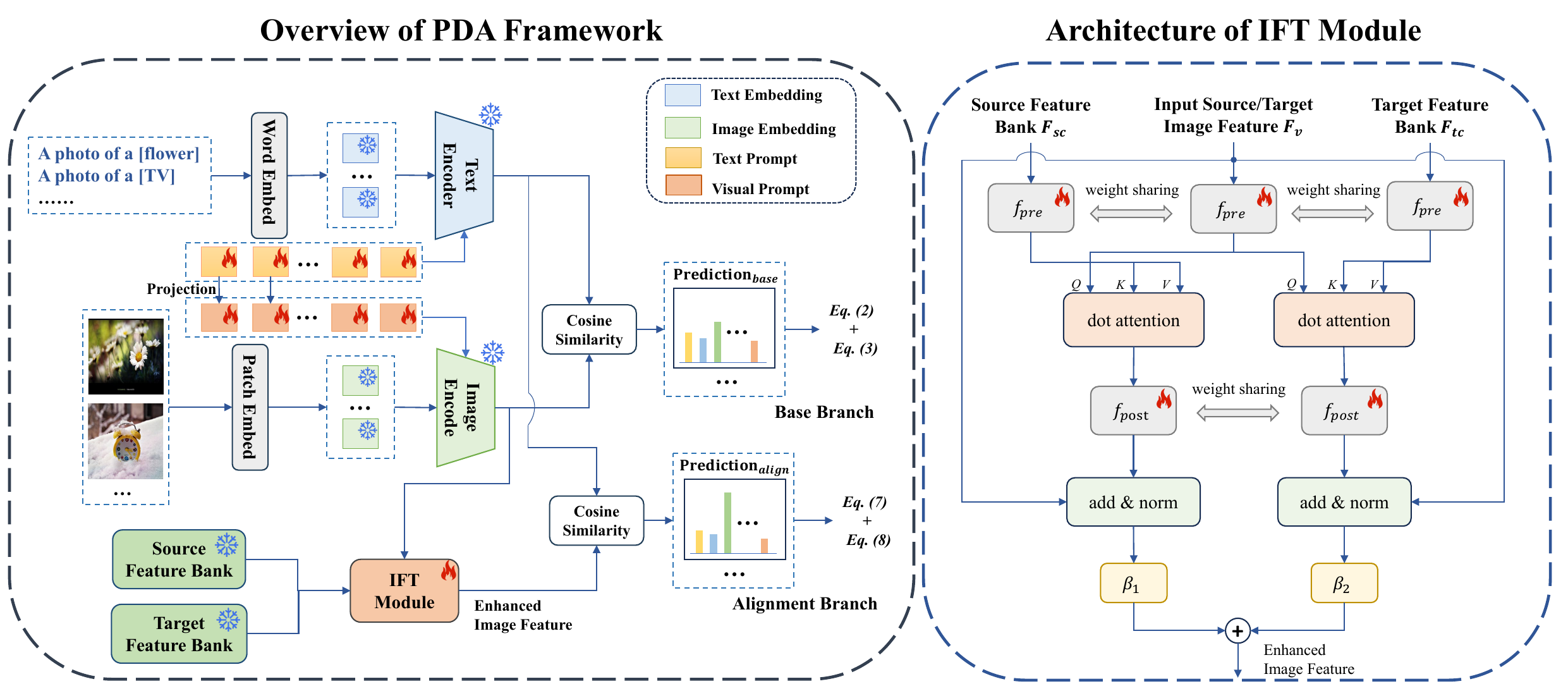}
  \caption{Overview of the proposed Prompt-based Distribution Alignment (PDA) method. The \textcolor{blue}{snow} denotes the frozen parameters, and the \textcolor{red}{fire} denotes the learnable parameters. 
  From left to right, we respectively show the detailed framework of PDA and the architecture of the IFT module.
  We mainly adopt the multi-modal prompt tuning in our PDA method. Additionally, IFT module makes the visual features attend to the source/target-domain feature bank for domain alignment. 
  }
  \label{fig:model}
\end{figure*}

Inspired by the observations in the previous section, we attempt to design an efficient yet effective prompt tuning method for UDA. To enhance the transferability of the prompt, we propose a Prompt-based Distribution Alignment (PDA) method, whose framework is illustrated in Figure \ref{fig:model}.
We introduce our PDA method as follows.

\subsection{Prompting for Base Branch}

\textbf{Prompt design.} We mainly adopt the paradigm of multi-modal prompt. 
For the early layers of the image encoder, a text prompt is employed to generate a visual prompt by a projection layer. This means that text prompts are employed to guide the encoding process of images, enabling the images to possess information in the feature space that is relevant to the given text, therefore achieving alignment of images with pertinent textual information.
For the later layers of the image encoder, each layer utilizes an independent prompt.  
This design allows each layer to independently capture distinct visual and semantic features of the image, enabling better image-text interaction and capturing different visual and semantic features.

\noindent \textbf{Loss function.} Contrastive loss function is then employed to align the image and text representations, which can be formulated as:

\begin{equation}\begin{aligned}\label{func:loss_x}
\mathcal{L}_x = -\sum_{i}^{}y^s_ilog\frac
{\exp (sim(\hat{w}_i, \hat{z}^s ) / t )}
{\sum^K_{j=1} \exp (sim(\hat{w}_j, \hat{z}^s ) / t )},
\end{aligned}\end{equation}
where $y^s$ denotes the one-hot ground-truth of source domain data, $K$ is the number of classes, $w_i$ and $\hat{z}^s$ denote the i-th class of final text representation and final image representations of the source domain with prompt tuning, respectively.

To further exploit data of the target domain, we use pseudo labels to train these unlabeled data like Ge et al.~\cite{ge2022domain}. The pseudo labels are generated by the prediction of CLIP model. In order to enhance the reliability of these pseudo labels, we set a fixed threshold value $\tau$. If the maximum probability $\tau_p$ predicted by CLIP for a given image is lower than this threshold, the pseudo label is discarded.
Again, we adopt the contrastive loss function:

\begin{equation}\begin{aligned}\label{func:loss_u}
\mathcal{L}_u = -\mathbb{I}{(\tau_p\geq\tau)}\sum_{i}^{}\hat{y}^t_ilog\frac
{\exp (sim(\hat{w}_i, \hat{z}^t ) / t)}
{\sum^K_{j=1} \exp (sim(\hat{w}_j, \hat{z}^t ) / t)},
\end{aligned}\end{equation}
where $\mathbb{I}(\cdot)$ is an indicator function, $\hat{y}^t$ denotes the one-hot pseudo label of target domain data and $\hat{z}^t$ denotes final image representations of the target domain with prompt tuning.

\subsection{Pipeline of Alignment Branch}

For the alignment branch, we construct feature banks for both the source and target domains and propose image-guided feature tuning (IFT) to make the input attend to feature banks to achieve domain alignment.

\noindent \textbf{Constructing feature banks.}  With access to data from both the source and target domains, we can obtain text features and image features from both domains. 
Based on the strong zero-shot ability of CLIP, we could construct robust and accurate feature banks. 
Firstly, we produce confidence scores (\textit{i.e}, maximum probability) for images in the source domain with the prediction in zero-shot CLIP. 
Similarly, we generate a confidence score and corresponding pseudo label for each image in the target domain. Specifically, the index of the maximum confidence score is the pseudo label of the image. 
We select the visual features of images with top-$C$ confidence scores in each class for the source and target domains, and construct a $K$-way $C$-shot source-domain feature bank and target-domain feature bank, where $K$ denotes the number of classes and $C$ denotes the number of samples in each class. Then we obtain the centroid features of each class as the final source-domain feature bank $z_{sc}$ and target-domain feature bank $z_{tc}$, respectively.

\begin{table*}[ht]
\centering
\begin{tabular}{c|ccccccccccccc}
\toprule
Method & A-C & A-P & A-R & C-A & C-P & C-R & P-A & P-C & P-R & R-A & R-C & R-P & Avg \\
\midrule
zero-shot CLIP & 67.6 & 89.0 & 89.4 & 82.4 & 89.0 & 89.4 & 82.4 & 67.6 & 89.4 & 82.4 & 67.6 & 89.0 & 82.1 \\
linear probe CLIP & 60.1 & 73.7 & 80.9 & 66.4 & 76.4 & 76.8 & 63.4 & 61.0 & 82.3 & 74.7 & 64.8 & 88.3 & 72.4 \\
CoOp & 70.0 & 90.8 & 90.9 & 83.2 & 90.9 & 89.2 & 82.0 & 71.8 & 90.5 & 83.8 & 71.5 & 92.0 & 83.9 \\
CoCoOp & 70.4 & 91.4 & 90.4 & 83.5 & 91.8 & 90.3 & 83.4 & 70.9 & 91.0 & 83.4 & 71.2 & 91.7 & 84.1 \\
VP & 66.7 & 89.1 & 89.1 & 81.7 & 89.0 & 89.2 & 81.8 & 67.0 & 89.1 & 81.7 & 66.6 & 89.0 & 81.7 \\
VPT-shallow & 69.3 & 90.1 & 90.2 & 83.4 & 91.0 & 90.2 & 82.6 & 70.6 & 90.9 & 83.5 & 69.6 & 91.2 & 83.6 \\
VPT-deep & 71.6 & 89.9 & 90.3 & 82.8 & 91.0 & 89.7 & 82.0 & 71.5 & 90.3 & 84.6 & 71.7 & 91.6 & 83.9 \\
IVLP & 71.4 & \textbf{91.7} & 90.8 & 83.6 & 90.2 & 89.3 & 82.2 & 72.4 & 90.4 & 84.1 & 72.1 & 92.0 & 84.2 \\
MaPLe & 72.2 & 91.6 & 90.3 & 82.6 & 90.9 & 89.8 & 82.4 & 71.6 & 90.1 & 85.1 & 72.0 & 92.1 & 84.2 \\
DAPL & 70.7 & 91.0 & 90.9 & 85.2 & 91.0 & 91.0 & 85.1 & 70.7 & 90.9 & 85.3 & 70.4 & 91.4 & 84.4 \\
\rowcolor{gray!30} \textbf{PDA (Ours)} & \textbf{73.5} & 91.4 & \textbf{91.3} & \textbf{86.0} & \textbf{91.6} & \textbf{91.5} & \textbf{86.0} & \textbf{73.5} & \textbf{91.7} & \textbf{86.4} & \textbf{73.0} & \textbf{92.4} & \textbf{85.7} \\
\bottomrule
\end{tabular}
\caption{Comparisons with the prompt tuning methods on Office-Home dataset with ViT-B/16 as the backbone. Bold denotes the best scores.}
\label{tab:officehome-ViT}
\end{table*}

\begin{table}[ht]
\centering
\resizebox{0.47 \textwidth}{!}{
\begin{tabular}{c|ccccccc}
\toprule
Method & A-D & A-W & D-A & D-W & W-A & W-D & Avg \\
\midrule
zero-shot CLIP & 77.7 & 75.8 & 79.0 & 75.8 & 79.0 & 77.7 & 77.5 \\
linear probe CLIP & 83.1 & 83.3 & 74.2 & 96.5 & 70.3 & 98.4 & 84.3 \\
CoOp & 88.5 & 88.5 & 82.0 & 96.1 & 82.4 & 99.0 & 89.4 \\
CoCoOp & 86.9 & 88.2 & 83.2 & 94.1 & \textbf{82.8} & 98.0 & 88.9 \\
VP & 78.5 & 74.8 & 77.9 & 75.5 & 77.8 & 79.7 & 77.4 \\
VPT-shallow & 83.5 & 83.8 & 77.5 & 88.6 & 80.9 & 91.2 & 84.2 \\
VPT-deep & 89.6 & 86.5 & 81.9 & 96.5 & \textbf{82.8} & 99.2 & 89.4 \\
IVLP & 85.7 & 89.2 & 81.9 & \textbf{98.4} & 80.3 & 99.2 & 89.1 \\
MaPLe & 86.9 & 88.6 & 83.0 & 97.7 & 82.0 & 99.4 & 89.6 \\
DAPL & 81.7 & 80.3 & 81.2 & 81.8 & 81.0 & 81.3 & 81.2 \\
\rowcolor{gray!30} \textbf{PDA (Ours)} & \textbf{91.2} & \textbf{92.1} & \textbf{83.5} & 98.1 & 82.5 & \textbf{99.8} & \textbf{91.2} \\
\bottomrule
\end{tabular}
}
\caption{Comparisons with the prompt tuning methods on Office-31 dataset with ViT-B/16 as the backbone. Bold denotes the best scores.}
\label{tab:office31-ViT}
\end{table}

\noindent \textbf{Image-guided feature tuning (IFT).} IFT leverages feature banks to guide images to obtain self-enhanced and cross-domain features, as shown in Figure \ref{fig:model} (right). 
We first apply a weight-shared projector layer $f_{pre}$, \textit{i.e.}, a three-layer multilayer perceptron, to transform the image feature $\hat{z}$, source-domain feature bank $z_{sc}$, and target-domain feature bank $z_{tc}$ into query, key and value, which can be formulated as:

\begin{equation}\begin{aligned}\label{func:trans}
Q = f_{pre}(\hat{z}), ~~ K_{sc},V_{sc} = f_{pre}(z_{sc}), \\
K_{tc}, V_{tc} = f_{pre}(z_{tc}).
\end{aligned}\end{equation}

We make the image feature attend to source-domain and target-domain feature banks, resulting in augmented image features. These features are then transformed by another weight-shared projector $f_{post}$. 
The whole process with attention can be formulated as:

\begin{equation}\begin{aligned}\label{func:attn_s}
z_{sa} = f_{post}(softmax(\frac{Q K^T_{sc}}{\epsilon})V_{sc}), \\
z_{ta} = f_{post}(softmax(\frac{Q K^T_{tc}}{\epsilon})V_{tc}),
\end{aligned}\end{equation}
where $\epsilon$ denotes the scale value and $T$ denotes the transpose operation.
Then, we combine an add and norm module with the original visual feature, which can be formulated as:

\begin{equation}\begin{aligned}\label{func:a_s}
z_{vs} = \frac{z_{sa}+\hat{z}}{||z_{sa}+\hat{z}||_2}, \\
z_{vt} = \frac{z_{ta}+\hat{z}}{||z_{ta}+\hat{z}||_2},
\end{aligned}\end{equation}
where $||\cdot||_2$ denotes 2-norm. Then the final augmented image representation $\hat{z}$ can be denoted as $\beta_1 z_{vs} + \beta_2 z_{vt}$.

\noindent \textbf{Loss function.} Contrastive loss function is then employed to align the image representations and feature banks of source and target domains, which can be formulated as:

\begin{equation}\begin{aligned}\label{func:loss_xa}
\mathcal{L}_{xa} = -\sum_{i}^{}y^s_ilog\frac
{\exp (sim(\hat{w}_i, h(\hat{z}^s) ) / t)}
{\sum^K_{j=1} \exp (sim(\hat{w}_j, h(\hat{z}^s) ) / t)},
\end{aligned}\end{equation}
where $h$ denotes the IFT module and $h(\hat{z}^s)$ denotes augmented image representations of the source domain.

Similar to the base branch, we use the data of the target domain and obtain augmented image representations of the target domain $\hat{z}^t$. Then contrastive loss function is adopted:

\begin{equation}\begin{aligned}\label{func:loss_ua}
\mathcal{L}_{ua} = -\mathbb{I}{(\tau_p\geq\tau)}\sum_{i}^{}\hat{y}^t_ilog\frac
{\exp (sim(\hat{w}_i, h(\hat{z}^t) ) / t)}
{\sum^K_{j=1} \exp (sim(\hat{w}_j, h(\hat{z}^t) ) / t)}.
\end{aligned}\end{equation}

As a result, our PDA method can be trained end-to-end using a total contrastive loss:

\begin{equation}\begin{aligned}\label{func:attnu}
\mathcal{L} = \mathcal{L}_x + \mathcal{L}_u + \gamma(\mathcal{L}_{xa} + \mathcal{L}_{ua}),
\end{aligned}\end{equation}
where $\gamma$ is hyper-parameter. During the test phase, we calculate a weighted sum of the predictions from both the base and alignment branches, resulting in the final prediction of our model. 
These two branches are essential not only for enhancing model discriminability but also for aligning the distribution shift between source and target domains.

\section{Experiments}
\label{sec:exp}

In the following section, we describe the datasets, baselines, experimental setup, and results of our analysis. Here we show essential comparison and analysis. More details and experiments are provided in the Appendix.

\subsection{Experimental Setting}

\begin{table*}[ht]
\centering
\resizebox{0.95 \textwidth}{!}{
\begin{tabular}{c|c|ccccccccccccc}
\toprule
Method & Backbone & A-C & A-P & A-R & C-A & C-P & C-R & P-A & P-C & P-R & R-A & R-C & R-P & Avg \\
\midrule
ERM & \multirow{7}{*}{{RN50}} & 34.9 & 50.0 & 58.0 & 37.4 & 41.9 & 46.2 & 38.5 & 31.2 & 60.4 & 53.9 & 41.2 & 59.9 & 46.1 \\
DANN &  & 45.6 & 59.3 & 70.1 & 47.0 & 58.5 & 60.9 & 46.1 & 43.7 & 68.5 & 63.2 & 51.8 & 76.8 & 57.6 \\
JAN &  & 45.9 & 61.2 & 68.9 & 50.4 & 59.7 & 61.0 & 45.8 & 43.4 & 70.3 & 63.9 & 52.4 & 76.8 & 58.3 \\
MDD &  & 54.9 & 73.7 & 77.8 & 60.0 & 71.4 & 71.8 & 61.2 & 53.6 & 78.1 & 72.5 & 60.2 & 82.3 & 68.1 \\
SHOT &  & \textbf{57.1} & 78.1 & 81.5 & 68.0 & 78.2 & 78.1 & 67.4 & 54.9 & 82.2 & 73.3 & \textbf{58.8} & 84.3 & 71.8\\
CDAN w/ SDAT &  & 56.0 & 72.2 & 78.6 & 62.5 & 73.2 & 71.8 & 62.1 & 55.9 & 80.3 & 75.0 & 61.4 & 84.5 & 69.5 \\
\rowcolor{gray!30} \textbf{PDA~(Ours)} &  & 55.4 & \textbf{85.1} & \textbf{85.8} & \textbf{75.2} & \textbf{85.2} & \textbf{85.2} & \textbf{74.2} & \textbf{55.2} & \textbf{85.8} & \textbf{74.7} & 55.8 & \textbf{86.3} & \textbf{75.3} \\

\midrule
TVT & \multirow{6}{*}{{ViT}} & 74.9 & 86.8 & 89.5 & 82.8 & 87.9 & 88.3 & 79.8 & 71.9 & 90.1 & 85.5 & 74.6 & 90.6 & 83.6 \\
SSRT &  & \textbf{75.2} & 89.0 & 91.1 & 85.1 & 88.3 & 89.9 & 85.0 & \textbf{74.2} & 91.2 & 85.7 & \textbf{78.6} & 91.8 & 85.4 \\
Deit-based & & 61.8 & 79.5 & 84.3 & 75.4 & 78.8 & 81.2 & 72.8 & 55.7 & 84.4 & 78.3 & 59.3 & 86.0 & 74.8 \\
CDTrans-Deit &  & 68.8 & 85.0 & 86.9 & 81.5 & 87.1 & 87.3 & 79.6 & 63.3 & 88.2 & 82.0 & 66.0 & 90.6 & 80.5 \\
CDAN w/ SDAT &  & 69.1 & 86.6 & 88.9 & 81.9 & 86.2 & 88.0 & 81.0 & 66.7 & 89.7 & 86.2 & 72.1 & 91.9 & 82.4 \\
\rowcolor{gray!30} \textbf{PDA~(Ours)} &  & 73.5 & \textbf{91.4} & \textbf{91.3} & \textbf{86.0} & \textbf{91.6} & \textbf{91.5} & \textbf{86.0} & 73.5 & \textbf{91.7} & \textbf{86.4} & 73.0 & \textbf{92.4} & \textbf{85.7} \\
\bottomrule
\end{tabular}
}
\vspace{-2mm}
\caption{Comparisons with SOTA methods on Office-Home with ResNet50 and ViT as the backbone. Bold is the best scores.}
\label{tab:officehome-sota}
\end{table*}

\begin{table*}[ht]
\centering
\resizebox{\textwidth}{!}{
\begin{tabular}{c|c|ccccccccccccc}
\toprule
Method & Backbone & plane & bicycle & bus & car & horse & knife & mcycl & person & plant & sktbrd & train & truck & Avg \\
\midrule
ERM & \multirow{7}{*}{{RN101}} & 55.1 & 53.3 & 61.9 & 59.1 & 80.6 & 17.9 & 79.7 & 31.2 & 81.0 & 26.5 & 73.5 & 8.5 & 52.4 \\
DANN &  & 81.9 & 77.7 & 82.8 & 44.3 & 81.2 & 29.5 & 65.1 & 28.6 & 51.9 & 54.6 & 82.8 & 7.8 & 57.4 \\
MCD &  & 87.0 & 60.9 & 83.7 & 64.0 & 88.9 & 79.6 & 84.7 & 76.9 & 88.6 & 40.3 & 83.0 & 25.8 & 71.9 \\
MCC &  & 88.1 & 80.3 & 80.5 & 71.5 & 90.1 & \textbf{93.2} & 85.0 & 71.6 & 89.4 & 73.8 & 85.0 & 36.9 & 78.8 \\
SHOT &  & 94.3 & \textbf{88.5} & 80.1 & 57.3 & 93.1 & 94.9 & 80.7 & 80.3 & 91.5 & 89.1 & 86.3 & 58.2 & 82.9 \\
CDAN w/ SDAT &  & 94.8 & 77.1 & 82.8 & 60.9 & 92.3 & 95.2 & 91.7 & \textbf{79.9} & \textbf{89.9} & \textbf{91.2} & 88.5 & 41.2 & 82.1 \\
\rowcolor{gray!30} \textbf{PDA (Ours)} &  & \textbf{97.2} & 82.3 & \textbf{89.4} & \textbf{76.0} & \textbf{97.4} & 87.5 & \textbf{95.8} & 79.6 & 87.2 & 89.0 & \textbf{93.3} & \textbf{62.1} & \textbf{86.4} \\

\midrule
TVT & {\multirow{6}{*}{ViT}} & 92.9 & 85.6 & 77.5 & 60.5 & 93.6 & 98.2 & 89.3 & 76.4 & 93.6 & 92.0 & 91.7 & 55.7 & 83.9 \\
SSRT &  & 98.9 & 87.6 & 89.1 & \textbf{84.8} & 98.3 & 98.7 & 96.3 & 81.1 & 94.8 & \textbf{97.9} & 94.5 & 43.1 & 88.8 \\
Deit-based &  & 98.2 & 73.0 & 82.5 & 62.0 & 97.3 & 63.5 & \textbf{96.5} & 29.8 & 68.7 & 86.7 & \textbf{96.7} & 23.6 & 73.2 \\
CDTrans-Deit &  & 97.1 & 90.5 & 82.4 & 77.5 & 96.6 & 96.1 & 93.6 & \textbf{88.6} & \textbf{97.9} & 86.9 & 90.3 & 62.8 & 88.4 \\
CDAN w/ SDAT &  & 96.3 & 80.7 & 74.5 & 65.4 & 95.8 & \textbf{99.5} & 92.0 & 83.7 & 93.6 & 88.9 & 85.8 & 57.2 & 84.5 \\
\rowcolor{gray!30} \textbf{PDA~(Ours)} &  & \textbf{99.2} & \textbf{91.1} & \textbf{91.9} & 77.1 & \textbf{98.4} & 93.6 & 95.1 & 84.9 & 87.2 & 97.3 & 95.3 & \textbf{65.3} & \textbf{89.7} \\
\bottomrule
\end{tabular}
}
\vspace{-2mm}
\caption{Comparisons with SOTA methods on VisDA-2017 with ResNet101 and ViT as the backbone. Bold is the best scores.}
\label{tab:visda17-sota}
\end{table*}

\textbf{Datasets.} 
Experiments are conducted on popular benchmark datasets of unsupervised domain adaptation, namely Office-Home~\cite{venkateswara2017deep}, Office-31~\cite{saenko2010adapting} and VisDA-2017~\cite{peng2018visda}.

\noindent \textbf{Baselines.} 
For prompt tuning methods, we choose 7 baselines, \textit{i.e.,} CoOp~\cite{zhou2022learning}, CoCoOp~\cite{zhou2022conditional}, VPT~\cite{jia2022visual}, VP~\cite{bahng2022exploring}, IVLP~\cite{khattak2023maple}, MaPLe~\cite{khattak2023maple} and DAPL~\cite{ge2022domain}. 
We also compare PDA with the state-of-the-art (SOTA) methods, including ResNet-based and ViT-based methods. The ResNet-based methods are DANN~\cite{ganin2015unsupervised}, JAN~\cite{long2017deep}, MCD~\cite{saito2018maximum}, MDD~\cite{zhang2019bridging}, MCC~\cite{jin2020minimum}, SHOT~\cite{liang2020we} and SDAT~\cite{rangwani2022closer}, and the ViT-based methods are Deit~\cite{touvron2021training}, CDTrans~\cite{xu2021cdtrans}, SDAT, SSRT~\cite{sun2022safe} and TVT~\cite{yang2023tvt}.

\noindent \textbf{Experimental Setup.} 
We adopt ResNet50 ~\cite{he2016deep}, ResNet101 and ViT-B/16 ~\cite{dosovitskiy2020image} as our backbones. 
Following Zhou et al.~\cite{zhou2022learning}, we adopt text prompt as prompt design for ResNet-based backbone. Following Khattak et al.~\cite{khattak2023maple}, we adopt the multi-modal prompt as prompt design for the ViT-based backbone.
The parameters in the encoders of CLIP are fixed, and we train the prompt and IFT module using the SGD optimizer for 10 epochs on the Office-Home and VisDA-2017 datasets, and for 20 epochs on the Office-31 dataset, with a batch size of 32. 
For all prompt tuning methods, we set the learning rate initially to around 0.003 initially and decay it using a cosine annealing rule. Moreover, the context tokens length is set to 2 for MaPLe and our PDA method, 10 for VPT and VP, and 16 for CoOp and CoCoOp. 

\subsection{Comparisons with Prompt Tuning Methods}

\noindent\textbf{Results on Office-Home.} As shown in Table \ref{tab:officehome-ViT}, our PDA achieves the best performance on almost all tasks with 85.7\% accuracy, and achieves an average accuracy improvement of 3.6\%, 1.8\%, and 1.5\%, respectively, compared with zero-shot CLIP, CoOp and MaPLe. For some tasks, such as C-A and P-A, we observe improvements of around 4.0\% compared with MaPLe. Furthermore, we find that multi-modal prompt tuning methods perform better than single-modal prompt tuning methods.

\begin{figure*}[htbp]
  \centering
  \includegraphics[width=\textwidth]{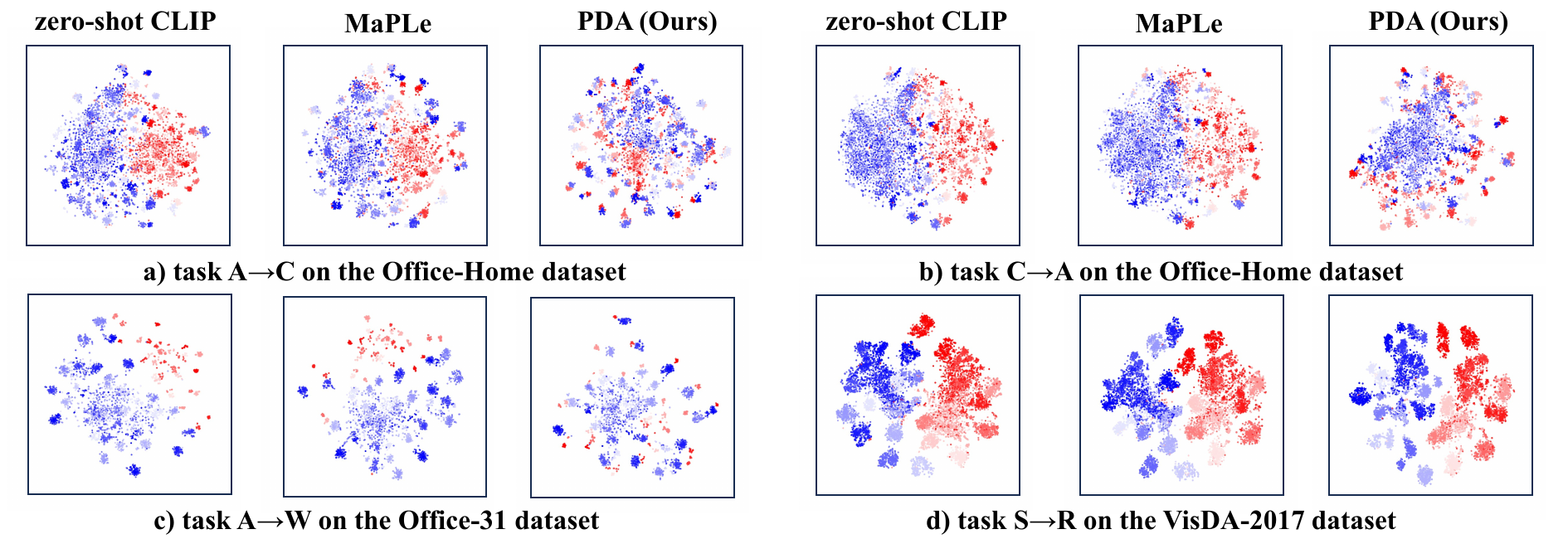}
  \caption{The t-SNE visualization for different tasks on the three datasets with zero-shot CLIP, MaPLe and our PDA method. Image features extracted from the source and target domain are shown in blue and red, respectively.
  }
  \label{fig:tsne}
\end{figure*}

\noindent \textbf{Results on Office-31.} 
As shown in Table \ref{tab:office31-ViT}, our PDA method also outperforms all other prompt tuning methods. We observe that prompt tuning can significantly improve the transferability of zero-shot CLIP, as PDA outperforms zero-shot CLIP by 13.7\% on average accuracy. For some tasks, such as W-D and D-W, our PDA outperforms zero-shot CLIP by 22.1\% and 22.3\%, respectively, indicating that the domain shift problem is well alleviated. 

\subsection{Comparisons with SOTA Methods} 

\noindent \textbf{Results on Office-Home.}  
Table \ref{tab:officehome-sota} shows the quantitative comparison with the ResNet-based and ViT-based methods. PDA outperforms other SOTA methods with identical backbones.
For instance, with ResNet50 as the backbone, PDA outperforms SHOT by 3.5\% and SDAT by 5.8\% by a large margin, respectively. With ViT as the backbone, PDA outperforms SSRT by 0.3\% and TVT by 2.1\%,  respectively. Compared with these unimodal methods, PDA exhibits superior performance with multi-modal interaction.

\noindent \textbf{Results on VisDA-2017.} 
Table \ref{tab:visda17-sota} shows the experimental results on the VisDA-2017 dataset. Our PDA method also achieves SOTA performance on the VisDA-2017 dataset with different backbones. For example, PDA outperforms SHOT and SDAT by a large margin of 3.5\% and 4.3\%, respectively. With ResNet101 as the backbone, PDA outperforms SSRT by 1.1\% and TVT by 5.8\%, respectively.

\subsection{Ablation Study} 

\textbf{Effect of each constraint loss.} 
Table \ref{tab:abla_loss} shows the experimental results of integrating different constraint losses. In most cases, each constraint loss contributes positively to enhancing the model's performance. 
For Office-Home dataset, we observe a consistent performance improvement with the introduction of each constraint loss, and the combination of them improves the averaged results by 3.6\%. 
For Office-31 dataset, a notable improvement of 12.1\% is achieved by incorporating the $\mathcal{L}_x$, which ensures discrimination among different classes. The combined influence of these constraint losses results in an impressive average performance improvement of 13.7\%.
For VisDA-2017 dataset, we encounter a tendency towards overfitting to data of source domain when employing $\mathcal{L}_x$, but this issue is mitigated by the application of other constraint losses.

\begin{table}[ht]
\centering
\resizebox{0.47 \textwidth}{!}{
\begin{tabular}{cccc|ccc}
\toprule
$\mathcal{L}_x$ & $\mathcal{L}_{xa}$ & $\mathcal{L}_u$ & $\mathcal{L}_{ua}$ & OfficeHome & Office31 & VisDA17 \\
\midrule
~ & ~ & ~ & ~ & 82.1 & 77.5 & 88.9 \\
$\checkmark$ & ~ & ~ & ~ & 84.2 (\textcolor[cmyk]{1, 0, 1, 0.61}{+2.1}) & 89.6 (\textcolor[cmyk]{1, 0, 1, 0.61}{+12.1}) & 83.5 (\textcolor[cmyk]{0, 0.75, 0.75, 0.35}{-5.4}) \\
$\checkmark$ & $\checkmark$ & ~ & ~ & 84.6 (\textcolor[cmyk]{1, 0, 1, 0.61}{+2.5}) & 89.8 (\textcolor[cmyk]{1, 0, 1, 0.61}{+12.3}) & 85.2 (\textcolor[cmyk]{0, 0.75, 0.75, 0.35}{-3.7}) \\
$\checkmark$ & $\checkmark$ & $\checkmark$ & ~ & 85.2 (\textcolor[cmyk]{1, 0, 1, 0.61}{+3.1}) & 90.5 (\textcolor[cmyk]{1, 0, 1, 0.61}{+13.0}) & 89.0 (\textcolor[cmyk]{1, 0, 1, 0.61}{+0.1}) \\
$\checkmark$ & $\checkmark$ & $\checkmark$ & $\checkmark$ & \textbf{85.7 (\textcolor[cmyk]{1, 0, 1, 0.61}{+3.6})} & \textbf{91.2 (\textcolor[cmyk]{1, 0, 1, 0.61}{+13.7})} & \textbf{89.7 (\textcolor[cmyk]{1, 0, 1, 0.61}{+0.8})}  \\
\bottomrule
\end{tabular}
}
\caption{Ablation on different constraint losses. The average results of three datasets are reported. Improvements over the baseline of CLIP are marked in green.}
\label{tab:abla_loss}
\end{table}

\begin{figure}[ht]
  \centering
  \includegraphics[width=0.472\textwidth]{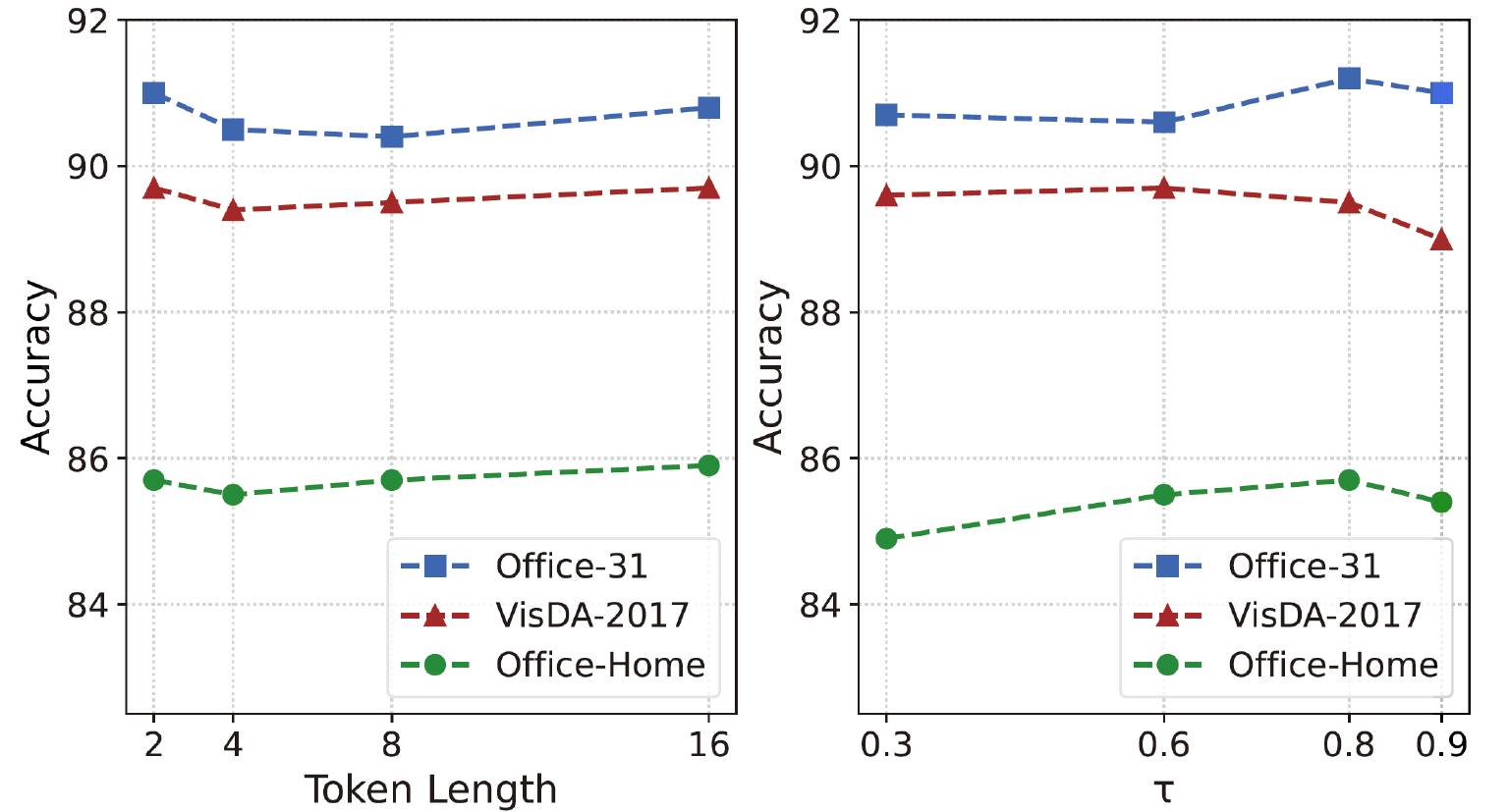}
  \caption{Sensitivity analysis of the context token length (left) and pseudo label threshold $\tau$ (right) on three datasets.}
  \label{fig:ablation}
\end{figure}

\noindent \textbf{Sensitivity analysis of the pseudo label threshold $\tau$ and context token length.} Figure \ref{fig:ablation} presents the results of varying the context token length and pseudo label threshold, respectively. The results suggest that the performance of our method is generally robust to both of them.

\subsection{Visualization}
As shown in Figure \ref{fig:tsne}, we visualize the image features extracted from zero-shot CLIP, MaPLe and our PDA on four tasks from the three datasets via t-SNE. We can observe that our PDA method can better align the two domains.

\section{Conclusion}
\label{sec:conclu}

In this paper, we demonstrate the effectiveness of vision language models and prompt tuning of VLMs for unsupervised domain adaptation.
Based on this, we introduce distribution alignment into prompt tuning and propose a Prompt-based Distribution Alignment (PDA) method with a two-branch training paradigm. 
These two branches play a vital role not only in improving model discriminability but also in mitigating the distribution shift between the source and target domains.
Extensive experiments confirm the effectiveness of our proposed method and our PDA method achieves new state-of-the-art performance for unsupervised domain adaptation. Due to the transferability of the learned prompts, we may further explore prompt alignment for unsupervised domain adaptation or other downstream tasks in future work.

\section*{Acknowledgments}
This work was supported by the National Natural Science Foundation of China under grant
number U21A20485, 62088102, and NSFC General Program under grant number 62176215.

\bibliography{aaai24}

\newpage
\newpage
\appendix


\begin{center}
\textbf{\Large Supplementary Material}
\end{center}

This appendix is organized as follows: 

\begin{itemize}

\item Section~\ref{sec:DD} provides the detailed dataset information.

\item Section~\ref{sec:ID} provides the detailed additional training implementation details.

\item Section~\ref{sec:SCE} gives additional experiment results, including additional comparison with prompt tuning methods and SOTA methods, additional visualization, and the analysis of the metrics that measure the performance.

\end{itemize}

\section{Dataset Details}
\label{sec:DD}

\noindent \textbf{Office-31} ~\cite{saenko2010adapting}. The Office-31 dataset is a popular small-scaled benchmark for domain adaptation. It consists of 4,110 images of 31 categories, with three distinct domains: Amazon (A), Webcam (W), and DSLR (D). In our experiment, we conduct 6 unsupervised domain adaptation (UDA) tasks, such as A $\rightarrow$ D, A $\rightarrow$ W, and so on.

\noindent \textbf{Office-Home} ~\cite{venkateswara2017deep}. The Office-Home dataset is a medium-scaled benchmark for domain adaptation. It contains a total of 15,500 images from four distinct domains: Art (A), Clip Art (C), Product (P), and Real World (R). Each domain contains objects from 65 categories commonly found in office and home environments. To evaluate the effectiveness of our proposed method, we conduct 12 unsupervised domain adaptation (UDA) tasks, which involve adapting models trained on a source domain to a target domain, for example, A $\rightarrow$ C, A $\rightarrow$ P, and so on.

\noindent \textbf{VisDA-2017} ~\cite{peng2018visda}. The Visual Domain Adaptation Challenge 2017 (VisDA17) dataset is a more challenging large-scaled benchmark for synthetic-to-real domain adaptation. It consists of 12 categories, with a total of 152,397 synthetic images generated by rendering 3D models from different angles and light conditions, and 55,388 real-world images collected from the Microsoft Common Objects in Context (MSCOCO) dataset. In our experiments, we follow the setting of ~\cite{ge2022domain} and treat the synthetic images as the source domain and the real-world images as the target domain, \textit{i.e.}, S $\rightarrow$ R.

\section{Implementation Details}
\label{sec:ID}

We adopt the pre-trained CLIP model with ResNet50, ResNet101 and ViT-B/16 as our backbones. 
For all prompt tuning methods, we fix the parameters in the encoders and the prompt is trained with the SGD optimizer for 10 epochs for Office-Home and VisDA-2017 datasets, 20 epochs for Office-31 dataset, where the batch size is 32. Referring to the original paper, the learning rate is set to 0.001 for CoCoOp, 0.0025 for VPT and VP, 0.003 for CoOp, DAPL and our PDA method, and 0.0035 for IVLP and MaPLe.
The pseudo labeling threshold $\tau$ is set to 0.8 for Office-Home, 0.8 for Office-31 and 0.6 for VisDA-2017 dataset. The weight $\gamma$ of constraint losses is set to 1.0. The weights $\beta_1, \beta_2$ of the enhanced augmented image features are set to 0.1 and 0.1. For N-way K-shot feature banks, N is set to the number of classes in each dataset, and K is set to 5. 

For the Office-Home and VisDA-2017 datasets, we utilize the logits of zero-shot CLIP to generate pseudo labels due to their high average accuracy with zero-shot CLIP. However, for the Office-31 dataset, the average accuracy of zero-shot CLIP is lower. Therefore, we apply a warm-up strategy, in which we use the logits of zero-shot CLIP for the first couple of epochs and leverage the logits of prompt tuning CLIP for the last couple of epochs. If the accuracy is still lower, we only use pseudo labels for the last few epochs.

\section{Supplement Comparison Experiments}
\label{sec:SCE}

The supplement experiments mainly demonstrate the effectiveness of our PDA method across different backbones. Furthermore, to give a more detailed analysis, we conduct an additional ablation study, additional visualization experiments, and a metric analysis of our PDA method.

\subsection{Comparisons with Prompt Tuning Methods}

\textbf{Results on Office-Home.} As shown in Table \ref{tab:officehome-rn}, we observe a stable improvement in performance across 12 tasks with different backbones on the Office-Home dataset, and our method achieves state-of-the-art performance among all prompt tuning methods.

\begin{table*}[htbp]
\centering
\resizebox{\textwidth}{!}{
\begin{tabular}{c|c|ccccccccccccc}
\toprule
Method & Backbone & A-C & A-P & A-R & C-A & C-P & C-R & P-A & P-C & P-R & R-A & R-C & R-P & Avg \\
\midrule
zero-shot CLIP & \multirow{7}{*}{RN50} & 51.7 & 81.5 & 82.3 & 71.7 & 81.5 & 82.3 & 71.7 & 51.7 & 82.3 & 71.7 & 51.7 & 81.5 & 71.8 \\
linear probe CLIP &  & 32.3 & 40.3 & 52.7 & 40.1 & 50.0 & 50.1 & 49.3 & 35.0 & 71.7 & 61.5 & 37.6 & 74.4 & 49.6 \\
CoOp &  & 53.8 & 84.6 & 84.1 & 71.0 & 83.1 & 82.3 & 70.4 & 53.8 & 83.6 & 74.2 & 54.6 & \textbf{87.0} & 73.5 \\
CoCoOp &  & 52.1 & 84.6 & 84.9 & 72.0 & 85.1 & 84.0 & 69.3 & 54.3 & 84.5 & 72.6 & 53.2 & 86.8 & 73.6 \\
VP &  & 51.8 & 81.6 & 82.6 & 71.9 & 81.7 & 82.5 & 71.8 & 51.0 & 82.6 & 72.0 & 51.5 & 81.7 & 71.9 \\
DAPL &  & 54.1 & 84.3 & 84.8 & 74.4 & 83.7 & 85.0 & 74.5 & 54.6 & 84.8 & 75.2 & 54.7 & 83.8 & 74.5 \\
\rowcolor{gray!30} \textbf{PDA (Ours)} &  & \textbf{55.4} & \textbf{85.1} & \textbf{85.8} & \textbf{75.2} & \textbf{85.2} & \textbf{85.2} & \textbf{74.2} & \textbf{55.2} & \textbf{85.8} & \textbf{74.7} & \textbf{55.8} & 86.3 & \textbf{75.3} \\

\midrule
zero-shot CLIP & \multirow{7}{*}{RN101} & 56.1 & 85.8 & 85.3 & 77.2 & 85.8 & 85.3 & 77.2 & 56.1 & 85.3 & 77.2 & 56.1 & 85.8 & 76.1 \\
linear probe CLIP &  & 38.7 & 47.4 & 57.5 & 52.1 & 60.5 & 59.7 & 55.8 & 45.9 & 76.1 & 69.0 & 48.2 & 78.7 & 57.5 \\
CoOp &  & 59.7 & 86.6 & 87.0 & 76.6 & 85.7 & 85.9 & 77.0 & \textbf{61.0} & 86.5 & 79.2 & 60.6 & 88.6 & 77.9 \\
CoCoOp &  & 59.1 & 86.6 & 86.8 & 77.0 & 87.7 & 86.2 & 76.3 & 60.2 & 87.1 & 79.4 & 59.8 & 88.4 & 77.9 \\
VP &  & 55.6 & 86.1 & 85.3 & 76.1 & 86.1 & 85.5 & 76.6 & 55.8 & 85.5 & 76.1 & 56.0 & 86.1 & 75.9 \\
DAPL &  & 59.0 & \textbf{88.1} & 87.5 & 79.6 & 87.9 & 87.1 & 79.6 & 58.8 & 87.2 & \textbf{79.8} & 58.8 & 88.0 & 78.5 \\
\rowcolor{gray!30} \textbf{PDA (Ours)} &  & \textbf{60.9} & 87.5 & \textbf{87.9} & \textbf{79.8} & \textbf{88.6} & \textbf{87.6} & \textbf{80.0} & 60.4 & \textbf{87.7} & 79.7 & \textbf{61.2} & \textbf{88.9} & \textbf{79.2} \\
\bottomrule
\end{tabular}
}
\caption{Comparisons with the prompt tuning methods. Top-1 classification accuracies on Office-Home dataset with ResNet50 and ResNet101 as the backbone. Bold denotes the best scores.}
\label{tab:officehome-rn}
\end{table*}

\begin{table*}[htbp]
\centering
\resizebox{\textwidth}{!}{
\begin{tabular}{c|c|ccccccccccccc}
\toprule
Method & Backbone & plane & bicycle & bus & car & horse & knife & mcycl & person & plant & sktbrd & train & truck & Avg \\
\midrule
zero-shot CLIP & \multirow{7}{*}{RN101} & 98.1 & 83.7 & 90.8 & 74.2 & 97.3 & 85.8 & 95.2 & 69.4 & 82.1 & \textbf{90.0} & 92.5 & 61.1 & 85.0 \\
linear probe CLIP &  & 77.1 & 38.2 & 63.0 & 72.7 & 80.4 & 10.3 & 97.7 & 13.6 & \textbf{89.5} & 70.5 & \textbf{96.1} & 4.2 & 59.4 \\
CoOp &  & 97.1 & \textbf{86.3} & 92.0 & 67.5 & 97.3 & 49.1 & 92.6 & 51.0 & 88.9 & 78.7 & 93.0 & 33.3 & 77.2 \\
CoCoOp &  & 96.5 & 75.0 & \textbf{94.1} & 58.4 & \textbf{98.3} & 59.3 & \textbf{96.1} & 62.9 & 88.4 & 85.5 & 88.6 & 67.0 & 80.8 \\
VP &  & \textbf{98.2} & 84.4 & 90.1 & 74.8 & 96.5 & 85.3 & 94.1 & 72.1 & 82.6 & 89.0 & 92.5 & 59.7 & 84.9 \\
DAPL &  & 98.1 & 83.9 & 90.9 & 75.1 & 97.6 & 87.1 & 95.2 & 75.7 & 84.8 & 90.7 & 91.8 & 61.7 & 86.1 \\
\rowcolor{gray!30} \textbf{PDA (Ours)} &  & 97.2 & 82.3 & 89.4 & \textbf{76.0} & 97.4 & \textbf{87.5} & 95.8 & \textbf{79.6} & 87.2 & 89.0 & 93.3 & \textbf{62.1} & \textbf{86.4} \\

\midrule
zero-shot CLIP & \multirow{11}{*}{ViT} & \textbf{99.2} & 92.2 & 93.5 & 76.7 & 98.3 & 90.4 & 94.6 & 83.6 & 85.4 & 96.1 & 94.3 & 62.5 & 88.9 \\
linear probe CLIP &  & 91.4 & 56.9 & 74.8 & 55.1 & 52.3 & 23.3 & 92.5 & 7.6 & 88.9 & 84.9 & 90.7 & 3.4 & 60.2 \\
CoOp &  & 98.7 & 89.8 & 94.2 & 69.7 & \textbf{99.0} & 71.5 & 96.3 & 53.9 & \textbf{91.5} & 96.3 & 95.8 & 35.7 & 82.7 \\
CoCoOp &  & 99.1 & 92.4 & 92.0 & 71.7 & 99.1 & \textbf{95.0} & 95.8 & 22.7 & 90.3 & 95.6 & 96.0 & 60.6 & 84.2 \\
VP &  & 99.0 & 91.2 & \textbf{93.8} & 77.0 & 98.3 & 89.1 & 94.4 & 85.7 & 82.9 & 95.1 & 94.1 & 63.8 & 88.7 \\
VPT-shallow &  & 99.0 & 86.8 & 95.7 & 69.3 & 98.5 & 73.0 & 96.5 & 78.3 & 80.3 & 96.2 & 93.4 & 56.1 & 85.3 \\
VPT-deep &  & 98.7 & 78.2 & 96.0 & 68.7 & 98.8 & 83.6 & \textbf{97.0} & 82.5 & 87.4 & 94.5 & 94.3 & 54.6 & 86.2 \\
IVLP &  & 98.6 & 86.8 & 88.5 & 76.0 & 97.5 & 68.3 & 95.7 & 59.0 & 90.5 & 94.5 & \textbf{97.9} & 36.3 & 82.5 \\
MaPLe &  & 98.6 & 85.8 & 93.0 & 68.8 & 99.2 & 72.4 & 96.8 & 77.1 & 84.7 & 96.0 & 95.9 & 33.1 & 83.5 \\
DAPL &  & 99.1 & \textbf{92.6} & 93.1 & \textbf{77.4} & 98.4 & 92.2 & 94.6 & 84.7 & 88.3 & 96.1 & 93.7 & 63.4 & 89.5 \\
\rowcolor{gray!30} \textbf{PDA (Ours)} &  & \textbf{99.2} & 91.1 & 91.9 & 77.1 & 98.4 & 93.6 & 95.1 & \textbf{84.9} & 87.2 & \textbf{97.3} & 95.3 & \textbf{65.3} & \textbf{89.7} \\
\bottomrule
\end{tabular}
}
\caption{Comparisons with the prompt tuning methods. Top-1 classification accuracies on VisDA-2017 dataset with ResNet101 and ViT-B/16 as the backbone. Bold denotes the best scores.}
\label{tab:visda17-rn}
\end{table*}

\begin{table*}[htbp]
\centering
\resizebox{0.6 \textwidth}{!}{
\begin{tabular}{c|c|ccccccc}
\toprule
Method & Backbone & A-D & A-W & D-A & D-W & W-A & W-D & Avg \\
\midrule
zero-shot CLIP & \multirow{7}{*}{RN50} & 74.1 & 67.0 & 72.8 & 67.0 & 72.8 & 74.1 & 71.3 \\
linear probe CLIP &  & 75.3 & 70.4 & 45.7 & 64.4 & 54.4 & 81.1 & 65.2 \\
CoOp &  & 82.3 & 78.2 & \textbf{77.9} & 90.7 & 76.3 & 96.4 & 83.6 \\
CoCoOp &  & 82.9 & 76.7 & 75.6 & 88.8 & 76.7 & 93.6 & 82.4 \\
VP &  & 74.9 & 68.4 & 73.9 & 68.4 & 74.1 & 76.1 & 72.6 \\
DAPL &  & 77.3 & 71.9 & 76.7 & 74.7 & \textbf{77.4} & 79.7 & 76.3 \\
\rowcolor{gray!30} \textbf{PDA (Ours)} &  & \textbf{85.1} & \textbf{81.1} & 76.6 & \textbf{92.8} & 77.3 & \textbf{97.8} & \textbf{85.1} \\

\midrule
zero-shot CLIP & \multirow{7}{*}{RN101} & 78.7 & 77.2 & 73.7 & 77.1 & 73.8 & 78.7 & 76.5 \\
linear probe CLIP &  & 79.7 & 77.6 & 49.8 & 66.7 & 59.2 & 86.1 & 69.9 \\
CoOp &  & 85.5 & 85.3 & 81.5 & 92.7 & 79.3 & \textbf{97.4} & 87.0 \\
CoCoOp &  & 84.5 & 83.6 & 80.5 & 93.1 & \textbf{80.3} & 96.8 & 86.5 \\
VP &  & 79.1 & 78.2 & 74.2 & 78.0 & 74.7 & 78.5 & 77.1 \\
DAPL &  & 83.9 & 81.0 & 78.1 & 83.5 & 78.6 & 82.7 & 81.3 \\
\rowcolor{gray!30} \textbf{PDA (Ours)} &  & \textbf{90.4} & \textbf{86.9} & \textbf{81.7} & \textbf{96.0} & 79.9 & \textbf{97.4} & \textbf{88.7} \\
\bottomrule
\end{tabular}
}
\caption{Comparisons with the prompt tuning methods. Top-1 classification accuracies on Office-31 dataset with ResNet50 and ResNet101 as the backbone. Bold denotes the best scores.}
\label{tab:office31-rn}
\end{table*}

\begin{table*}[ht]
\centering
\begin{tabularx}{\textwidth}
{cc>{\centering\arraybackslash}X>{\centering\arraybackslash}X>{\centering\arraybackslash}X>{\centering\arraybackslash}X>{\centering\arraybackslash}X>{\centering\arraybackslash}X}
\toprule
\multirow{3}{*}{Feature Type} 
    & Method & \multicolumn{2}{c}{zero-shot CLIP} & \multicolumn{2}{c}{MaPLe} & \multicolumn{2}{c}{PDA (Ours)} \\
\cmidrule(r){2-8} 
    & domain & source & target & source & target & source & target \\
\cmidrule(r){2-8} 
    & accuracy ($\uparrow$) & - & 67.6 & - & 72.2 & - & \textbf{73.5} \\
\midrule
\multirow{3}{*}{Text Feature} 
    & inner-class L2 distance ($\downarrow$) & \multicolumn{2}{c}{0.000} & \multicolumn{2}{c}{0.000} & \multicolumn{2}{c}{0.000} \\
    & inner-class variance ($\downarrow$) & \multicolumn{2}{c}{0.002} & \multicolumn{2}{c}{0.002} & \multicolumn{2}{c}{0.002} \\
    & \textbf{inter-class L2 distance} ($\uparrow$) & \multicolumn{2}{c}{0.685} & \multicolumn{2}{c}{0.595} & \multicolumn{2}{c}{\textbf{0.688}} \\
\midrule
\multirow{6}{*}{Image Feature} 
    & inner-class L2 distance ($\downarrow$) & 0.554 & 0.441 & 0.560 & 0.462 & \textbf{0.416} & \textbf{0.379} \\
    & inner-class variance ($\downarrow$) & 0.002 & 0.002 & 0.002 & 0.002 & 0.002 & 0.002 \\
    & inter-class L2 distance ($\uparrow$) & 0.717 & 0.625 & \textbf{0.744} & \textbf{0.717} & 0.605 & 0.627 \\
    & \textbf{r} ($\uparrow$) & 1.294 & 1.417 & 1.329 & 1.552 & \textbf{1.454} & \textbf{1.654} \\
    & \textbf{MMD} ($\downarrow$) & \multicolumn{2}{c}{0.307} & \multicolumn{2}{c}{0.260} & \multicolumn{2}{c}{\textbf{0.171}} \\
    & \textbf{KL divergence} ($\downarrow$) & \multicolumn{2}{c}{32.214} & \multicolumn{2}{c}{29.601} & \multicolumn{2}{c}{\textbf{28.517}} \\
\bottomrule
\end{tabularx}
\caption{Quantitative analysis of text and image features learned by zero-shot CLIP, MaPLe and PDA on Office-Home with ViT-B/16 backbone. The source domain is art, and the target domain is clipart. Bold denotes best scores.}
\label{tab:metric}
\end{table*}

\begin{table}[ht]
\centering
\resizebox{0.472 \textwidth}{!}{
\begin{tabular}{c|c|ccccccc}
\toprule
Method & & A-D & A-W & D-A & D-W & W-A & W-D & Avg \\
\midrule
ERM &  \multirow{6}{*}{\rotatebox{90}{RN50}} & 68.9 & 68.4 & 62.5 & 96.7 & 60.7 & 99.3 & 76.1 \\
DANN & & 79.7 & 82.0 & 68.2 & 96.9 & 67.4 & 99.1 & 82.2 \\
JAN & & 84.7 & 85.4 & 68.6 & 97.4 & 70.0 & 99.8 & 84.3 \\
MCD & & \textbf{92.2} & \textbf{88.6} & 69.5 & \textbf{98.5} & 69.7 & \textbf{100.0} & \textbf{86.5} \\
zero-shot CLIP &  & 74.1 & 67.0 & 72.8 & 67.0 & 72.8 & 74.1 & 71.3 \\
\rowcolor{gray!30} \textbf{PDA (Ours)} &  & 85.1 & 81.1 & \textbf{76.6} & 92.8 & \textbf{77.3} & 97.8 & \underline{85.1} \\

\midrule
Deit-based &  \multirow{4}{*}{\rotatebox{90}{ViT}} & 88.7 & 89.2 & 80.1 & 98.9 & 79.8 & 100.0 & 89.5 \\
CDTrans-Deit &  & \textbf{97.0} & \textbf{96.7} & 81.1 & \textbf{99.0} & 81.9 & \textbf{100.0} & \textbf{92.6} \\
zero-shot CLIP &  & 77.7 & 75.8 & 79.0 & 75.8 & 79.0 & 77.7 & 77.5 \\
\rowcolor{gray!30} \textbf{PDA (Ours)} &  & 91.2 & 92.1 & \textbf{83.5} & 98.1 & \textbf{82.5} & 99.8 & \underline{91.2} \\
\bottomrule
\end{tabular}
}
\caption{Comparisons with SOTA methods. Top-1 classification accuracies on Office-31 dataset with ResNet50 and ViT as the backbone. Bold denotes the best scores and underscore denotes the second-best result.}
\label{tab:office31-sota}
\end{table}

\noindent \textbf{Results on VisDA-2017.} Table \ref{tab:visda17-rn} shows the experimental results on the VisDA-2017 dataset and demonstrates PDA also achieves state-of-the-art performance among all prompt tuning methods.
Our observations indicate that most prompt tuning methods exhibit inferior performance compared to the zero-shot CLIP model, primarily due to the issue of overfitting. Interestingly, we also find an overfitting problem with the ResNet101 and Deit models, suggesting that the prompt has a similar characteristic to the model parameters.
Therefore, to learn an effective prompt, it is necessary to develop an appropriate training strategy to avoid overfitting problem.

\noindent \textbf{Results on Office-31.} Table \ref{tab:office31-rn} presents the experimental results on the Office-31 dataset. Our proposed PDA method significantly improves the transferability compared to zero-shot CLIP, and our PDA method achieves state-of-the-art performance among all prompt tuning methods.
The experimental results demonstrate the effectiveness of our proposed PDA method.

\subsection{Comparisons with SOTA Methods} 

Table \ref{tab:office31-sota} shows the experimental results on the Office-31 dataset. 
PDA outperforms most of the SOTA methods.
We observe that both the ResNet50 and Deit-based models outperform the zero-shot CLIP, suggesting that the CLIP-based model may be inadequate in handling relatively small domain shift problem. This highlights the efficacy of the conventional transfer learning approach employed by unimodal models in adapting more effectively to the data.
However, the prompt tuning methods can significantly improve the performance of CLIP model, enabling it to compare favorably with these state-of-the-art UDA methods with much fewer parameters.

\begin{figure}[ht]
  \centering
  \includegraphics[width=0.472\textwidth]{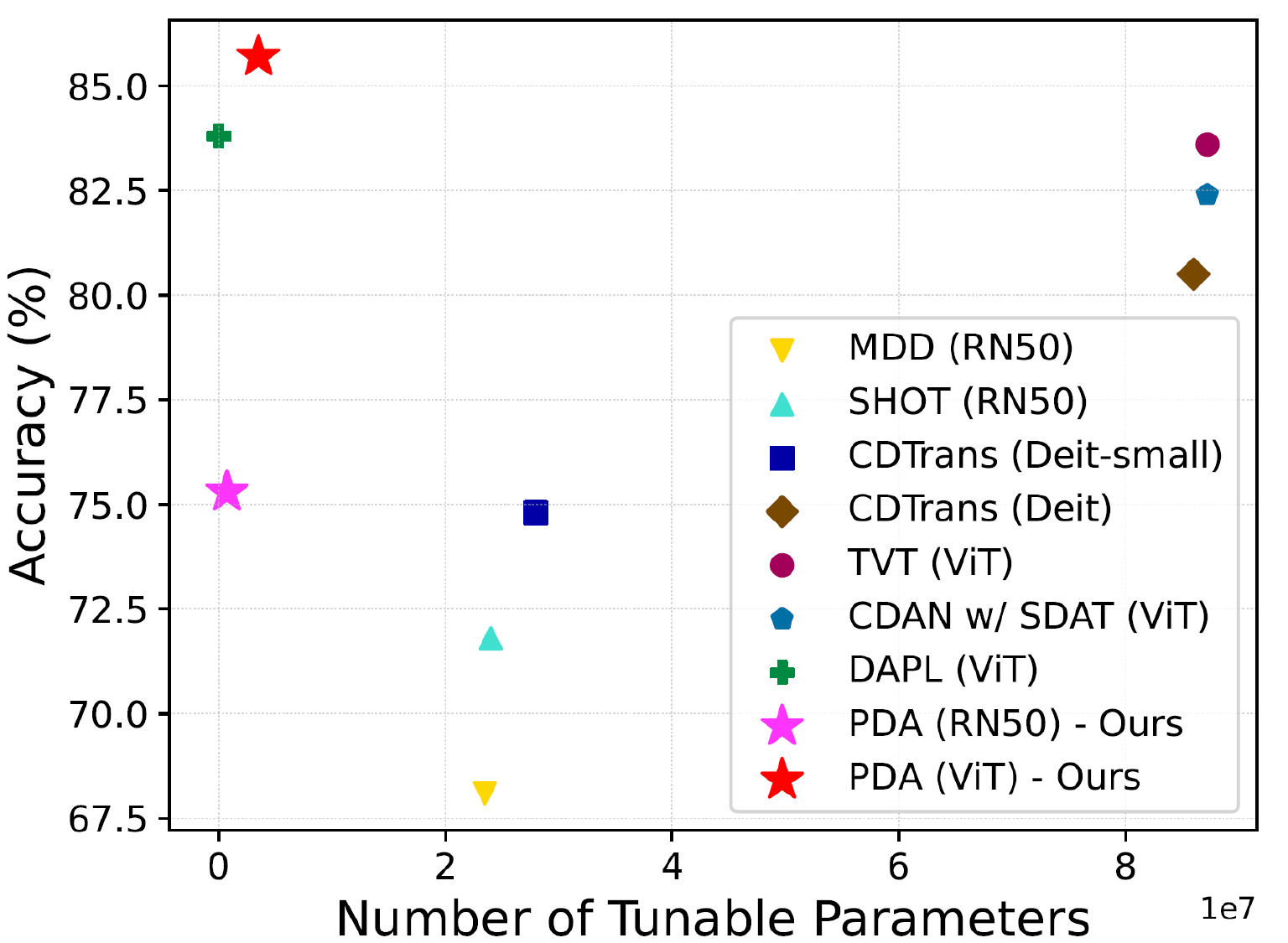}
  \caption{Performance comparison on Office-Home dataset. With much fewer parameters, our PDA method outperforms all other UDA methods}
  \label{fig:param}
\end{figure}

\subsection{Analyzing the Metrics}

Table \ref{tab:metric} summarizes the quantitative results of three methods. We evaluate the inner-class distance $D_1$ and variance, which are defined as the averaged $L_2$ distance and variance for inner-class features, respectively. Moreover, we evaluate the inter-class distance $D_2$, which is the average $L_2$ distance between the learned representations of each class and the centroid of other classes. We also define a ratio $r$ between the average of the inter-class distance and the inner-class distance, i.e., $r = D_2/D_1$, which denotes the compactness of the representations in each class.

\begin{table}[ht]
\centering
\resizebox{0.472 \textwidth}{!}{
\begin{tabular}{c|cccc|cccc}
\toprule
 & \multicolumn{4}{c|}{$\beta_s$=0.01} & \multicolumn{4}{c}{$\beta_s$=0.1} \\
\midrule
$\beta_t$ & 0.01 & 0.1 & 1 & 10 & 0.01 & 0.1 & 1 & 10 \\
\midrule
Acc & 85.4 & \textbf{85.8} & 85.6 & 81.1 & \underline{85.7} & \underline{85.7} & 85.4 & 81.0 \\
\midrule
& \multicolumn{4}{c|}{$\beta_s$=1} & \multicolumn{4}{c}{$\beta_s$=10} \\
\midrule
$\beta_t$ & 0.01 & 0.1 & 1 & 10 & 0.01 & 0.1 & 1 & 10 \\
\midrule
Acc & 85.4 & 85.3 & 85.2 & 81.2 & 80.7 & 80.8 & 80.5 & 79.7 \\
\bottomrule
\end{tabular}
}
\caption{Ablation on the weight of the final augmented image features.}
\label{tab:ab}
\end{table}

Compared to zero-shot CLIP, although the text features of MaPLe become more indiscriminable, the image features become more compact and discriminable, presenting a distance trade-off trait between text and image features. 
It is important to note that MaPLe minimizes the distribution shift due to the lower maximum mean discrepancy (MMD) and KL divergence. 
Our PDA method further introduces distribution into the prompt, and achieves a lower MMD and KL divergence, indicating the domain discrepancy minimizes. Moreover, we observe that PDA not only achieves a higher inter-class $L_2$ distance of text features but also a higher $r$ of image features, indicating both text features and image features become more discriminable.

\subsection{Ablation Study}

\textbf{Sensitivity analysis of the weight of the final augmented image features.} As shown in Table \ref{tab:ab}, we observe that optimal results are achieved when $\beta_t$ is approximately 0.1 and $\beta_s$ is around 0.1. However, if the weights of the final augmented image feature become excessively large, it could potentially compromise the model's discriminative ability.

\subsection{Visualization}

\textbf{Parameters analysis.} As shown in Figure \ref{fig:param}, our PDA method achieves superior results with considerably fewer parameters for both the ResNet50-based model and the ViT-based model.

\begin{figure}[ht]
  \centering
  \includegraphics[width=0.3 \textwidth]{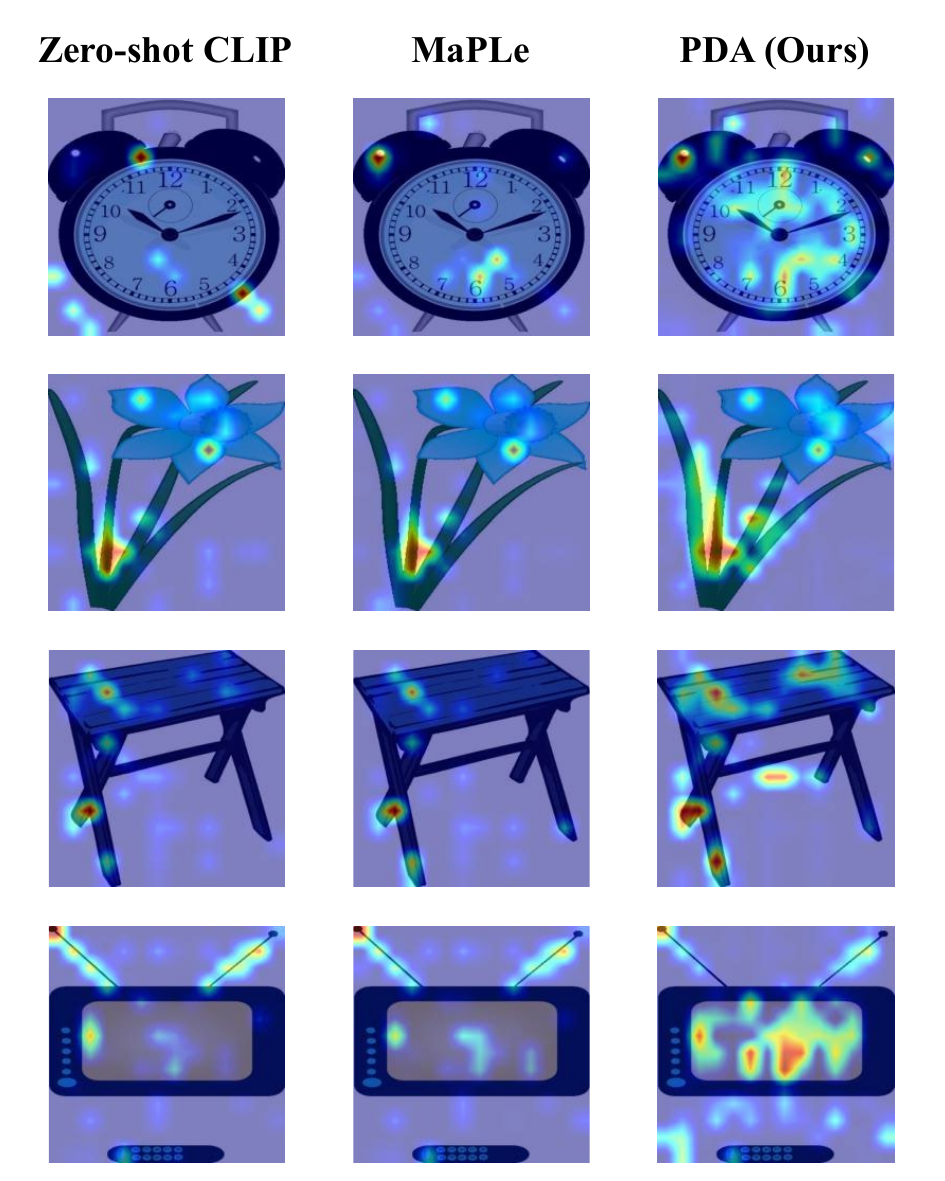}
  \caption{Attention visualization of zero-shot CLIP, MaPLe and PDA methods.}
  \label{fig:grad}
\end{figure}

\noindent \textbf{Attention.} In Figure \ref{fig:grad}, we visualize attention maps for different methods. We observe that the visual features of our method get more intensive and prominent.

\end{document}